\documentclass{amia}
\usepackage{graphicx}
\usepackage[labelfont=bf]{caption}
\usepackage[superscript,nomove]{cite}
\usepackage{color}
\usepackage{float}
\usepackage[symbol]{footmisc}
\usepackage{subcaption}
\usepackage{hyperref}

\begin{document}

\title{COVID-19 Smart Chatbot Prototype for Patient Monitoring}

\author{Hannah Lei$^{1}$\footnote[2]{\label{note1}Equal contributions}, Weiqi Lu$^{1}$\footref{note1} , Alan Ji$^{1}$, Emmett Bertram$^{1}$, Paul Gao$^{1}$,\\ Xiaoqian Jiang, Ph.D.$^{2}$, Arko Barman, Ph.D.$^{1}$\footnote{e-mail: arko.barman@rice.edu}}

\institutes{
    $^1$Rice University, Houston, Texas, United States;\\ $^2$The University of Texas Health Science Center at Houston\\
}

\maketitle

\noindent{\bf Abstract}

\textit{Many COVID-19 patients developed prolonged symptoms after the infection, including fatigue, delirium, and headache. The long-term health impact of these conditions is still not clear. It is necessary to develop a way to follow up with these patients for monitoring their health status to support timely intervention and treatment. In the lack of sufficient human resources to follow up with patients, we propose a novel smart chatbot solution backed with machine learning to collect information (i.e., generating digital diary) in a personalized manner. In this article, we describe the design framework and components of our prototype.
}

\section{Introduction}
COVID-19 has significantly changed how we seek medical assistance. During the pandemic, often patients will choose not to visit their doctor’s office as frequently as before due to the danger of infection. As a result, providers may not be informed of their conditions and symptoms in a timely manner\cite{kadambari_decrease_2020}. However, COVID-19 progresses quickly, and some newly-infected patients might develop severe symptoms if they do not see a doctor in time\cite{li_therapeutic_2020}. A chatbot will help build this bridge between doctors and patients self-isolating at home.

We developed a smart chatbot with the ability to collect symptoms and conditions about patients, prioritizing the known risk factors, and observe possible comorbidities (having another preexisting condition) with the advent of COVID-19. This chatbot also has the ability to hold follow-up conversations with patients to track the progression of symptoms and build a patient-specific profile using this information. 

\begin{figure}[t]
    \centering
    \includegraphics[width=1.0\linewidth, height=0.2\textheight]{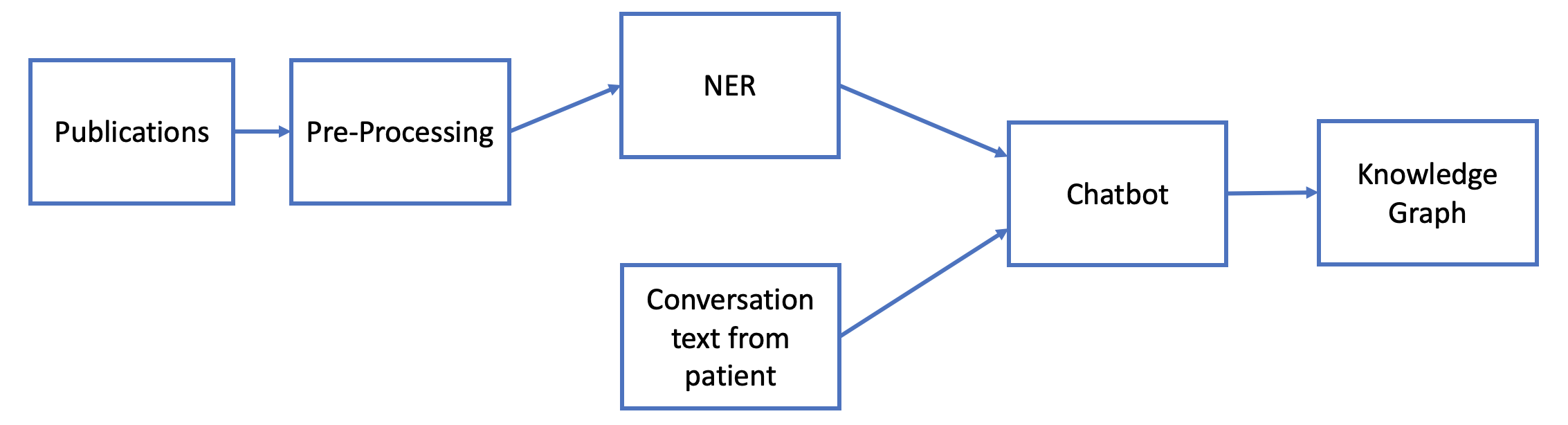}
    \caption{A block diagram of our method}
    \label{fig:block}
\end{figure}

To build the chatbot, we developed a natural language processing (NLP) pipeline which takes text data from published scientific literature as input and performs named entity recognition (NER) to extract names and associations of symptoms, drugs, and other relevant terminology. We then employed the information from the NER model to develop an end-to-end conversational chatbot. Combining the information from the patients with that from the NER model, we created a knowledge graph of symptoms which can be used for subsequent follow-up conversations with the patient. An overview of our model is shown in Figure~\ref{fig:block}.

\subsection{Contributions}

Our contributions in this work are as follows:

\begin{itemize}
    \item First, we developed a natural language processing (NLP) model using text data from the existing scientific research literature on COVID-19 and related diseases. This model performs named entity recognition (NER) to extract the names of symptoms, drugs, other terminology associated with the disease. These entities then form relationships and construct a knowledge graph for our chatbot to query.
    \item  Subsequently, we extended the pipeline to have the ability to extract patient information from conversational input, hold follow-up responses using the extracted data insights, and track the progression of patient symptoms and build a personalized profile.
\end{itemize}

Due to legal reasons, the chatbot cannot prescribe medicines and offer medical advice but instead directs the user or patient to a few associated sentences (or perhaps a summary) from a piece of literature in the dataset, and a link to an official guideline, after which they can follow up themselves with a doctor.

\section{Background and Related Work}

\subsection{Project Background}

The COVID‑19 pandemic is an ongoing pandemic of the novel coronavirus disease 2019 (COVID‑19). First identified in Wuhan, China, severe acute respiratory syndrome coronavirus 2 (SARS-CoV-2) is the virus responsible for the COVID-19. It has since been declared a Public Health Emergency of International Concern in January of 2020 and a pandemic in March of 2020~\cite{noauthor_01_nodate}. As of March 9, 2021, more than 118 million cases have been reported in 214 countries and territories, resulting in more than 2.5 million deaths\cite{dong_02_2020}.

The COVID-19 virus and pandemic have stimulated an upsurge in research, data sharing, and technological and medical innovation as the scientific world searches for solutions to combat and mitigate the disease. Further, these resources are used to track its spread and analyze the unfamiliar symptoms and effects the virus may have. While there are innumerable scientists from every country tackling this issue, the countermeasures for and long-term effects of COVID-19 will only be revealed with time. 

As mentioned, during the pandemic, many patients are opting not to go to the hospital and, thus, healthcare providers lack the opportunity to know their health conditions as quickly as possible. Doctors are using alternative strategies to check on their patients and limit hospital visits~\cite{PORTNOY20201489}. However, existing techniques like surveys or telemedicine have limitations~\cite{DAGGUBATI2020e859, sodhi_telehealth_2021}. The former can only focus on known risk factors while failing to collect other preexisting and life-threatening factors\cite{white_coronavirus_2020}. Telemedicine conducted by nurse practitioners has a bandwidth issue as individual nurses can only see a limited number of patients during the day\cite{sodhi_telehealth_2021}.

Our smart chatbot has a symptom tracking functionality allowing for a more comprehensive diagnosis when the patient has a chance to meet with a doctor. We use machine learning to help patients through intelligent conversations and collect information from potential patients, which is particularly valuable as hospital visits become more and more infrequent. Ultimately, this smart chatbot can potentially benefit the population at large, especially the elderly, who might have less inclination toward utilizing internet-based search engines to research their symptoms and risks. 

\begin{figure*}[t]
    \centering
    \subcaptionbox[]{Mobile environment\label{fig:mobile}}[.49\linewidth]{\includegraphics[width=0.6\linewidth, height=0.4\textheight]{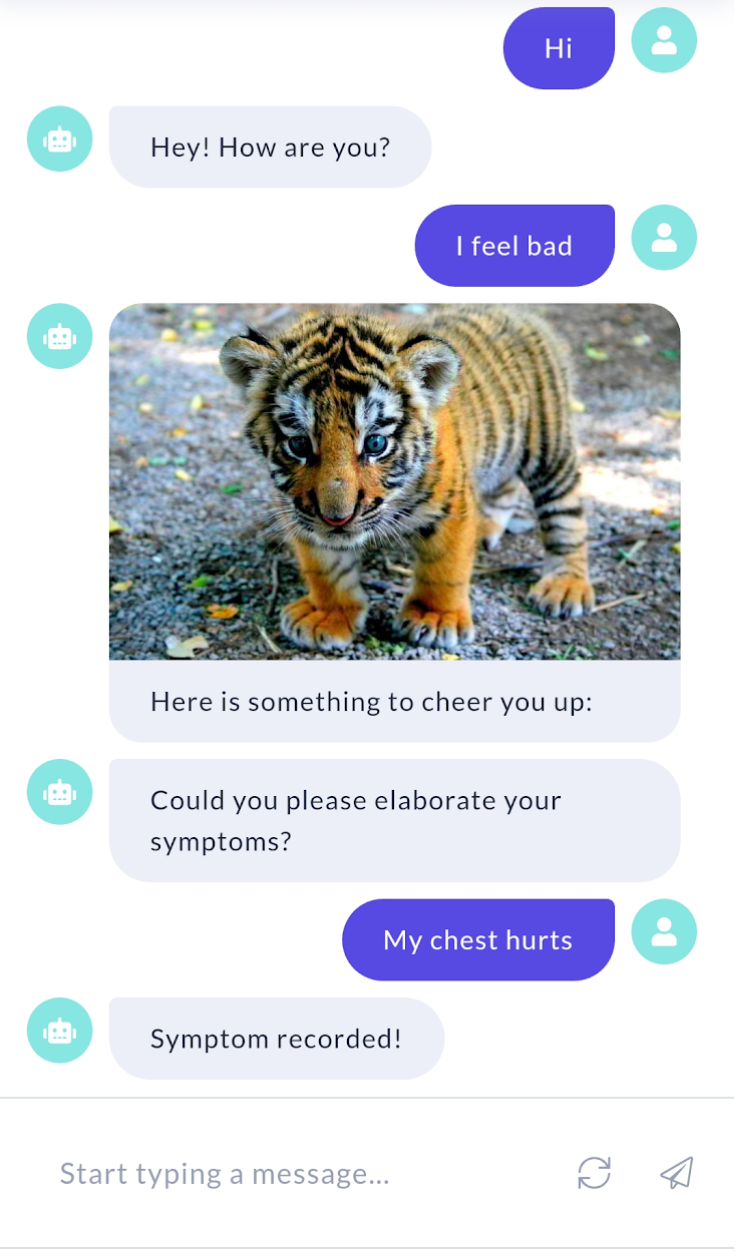}}
	\subcaptionbox[]{PC environment\label{fig:followup}}[.49\linewidth]{\includegraphics[width=\linewidth,height=0.22\textheight]{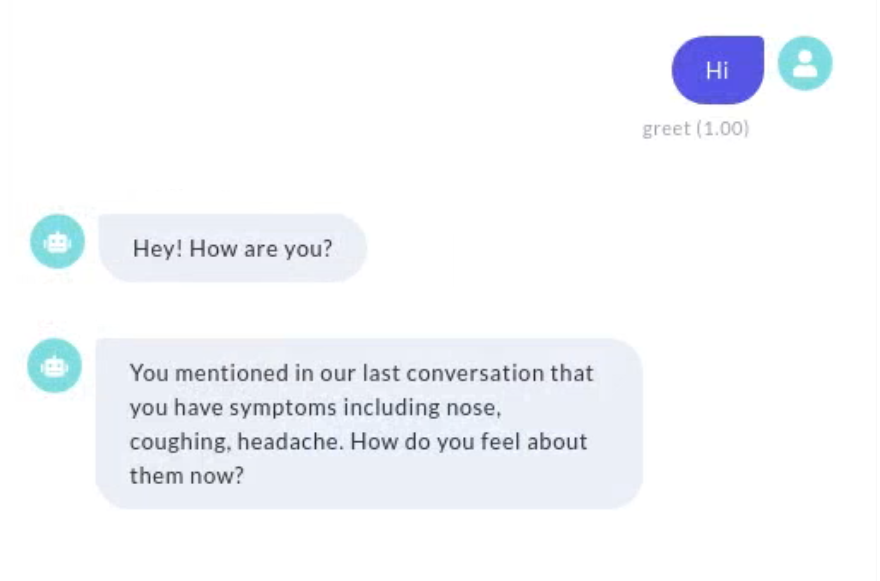}}
	\caption{A demo conversation between a patient and the chatbot in (a) mobile environment showing initial conversation, and (b) PC environment showing a follow-up conversation. The purple box is the patient input and grey box is the chatbot response.}
    \label{fig:demo}
\end{figure*}

Along with the essential features and functionalities of our chatbot, we also provide options for ``virtual appointment'' interactively and regularly. This allows the chatbot to occasionally check in on self-isolating or under-investigation patients for infection, monitor symptoms, and health conditions, and build a personalized patient profile, which will help healthcare providers make decisions based on evidence. A demo of our chatbot is shown in Figure~\ref{fig:demo}.

\begin{figure}[t]
    \centering
    \includegraphics[width=0.6\linewidth]{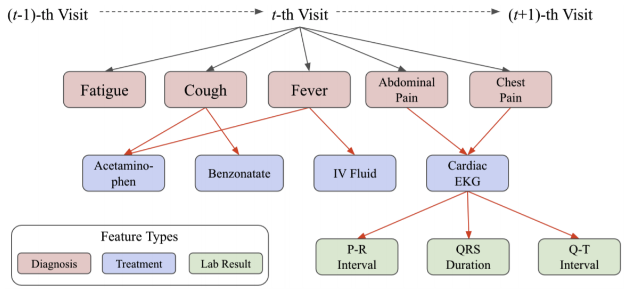}
    \caption{A knowledge graph representing a single patient’s visit to the doctor, with nodes related to the patient profile}
    \label{fig:kg}
\end{figure}

\subsection{Chatbots for COVID-19}

Penn Medicine, Google, and Verily worked in collaboration on a prior chatbot for answering questions about COVID-19 and doing basics assessment of symptoms\cite{herriman_asked_2020}. The application is targeted towards susceptible populations rather than confirmed COVID-19 patients and focused on a simple yes/no and categorical questions rather than performing NLU-based analysis on a patient’s input, such as descriptions of their symptoms. The user experience was measured directly by asking for feedback after giving answers. Another chatbot developed by researchers at the University of California, San Francisco~\cite{judson_implementation_2020} was designed for a screening of health system employees only and is not designed for use by the public.

\subsection{Named Entity Recognition}
In this work, we explored NER models specifically trained with a medical corpus using Med7\cite{kormilitzin_med7_2020}, which is a clinical NER with seven different categories: drug, form, route, frequency, dosage, strength, and duration (see Figure~\ref{fig:med7}). Some widely-used medical NER models rely on Bidirectional Encoder Representations from Transformers (BERT)\cite{devlin_bert_2019}. Medical models based on BERT are pre-trained with additional medical text corpora to capture domain-specific proper nouns and terms. BioBERT\cite{lee_biobert_2020}, ClinicalBERT\cite{alsentzer_publicly_2019}, and scispaCy\cite{scispacy} are two examples of such models that we explored in addition to Med7.

\subsection{Knowledge Graphs}
In order to track patients’ symptoms and activity over time, we constructed a knowledge graph by combining information about the patients gathered by the chatbot along with information from literature extracted using NER models. We store the information from a patient in a knowledge graph (an example is shown in Figure \ref{fig:kg}). The graph represents multiple different features, such as diagnosis, treatment, and lab results, as nodes. The edges in this particular graph represent the decision-making process of the physician. Typically, knowledge graphs are formed with directed weighted edges, where the weights express the existence of a semantic linkage between two entities\cite{manrique_knowledge_2018}.

\subsection{Rasa Natural Language Understanding (NLU)}
The Rasa Stack is a commonly used NLP library and tool specifically designed to aid the creation of chatbots. Rasa NLU is the approach that we used for intent classification and entity extraction from the patient input
\cite{liu_benchmarking_2019}. The Rasa NLU system has three components -- intent classification, entity extraction, and hyperparameter optimization. 





\begin{figure}[t]
    \centering
    \includegraphics[width=0.35\linewidth,height=0.1\textheight]{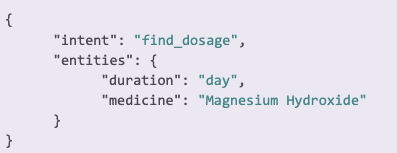}
    \caption{An example of an intent-entity structure returned by the Rasa stack.}
    \label{fig:intent}
\end{figure}

To showcase an example of the Rasa pipeline, we can formulate a query question, such as, \emph{“What is the dosage per day for my magnesium hydroxide prescription?”} With the Rasa stack, the sentence will be tokenized, and an intent-entity structure will be returned. An example is shown in Figure~\ref{fig:intent}), where the intent was classified as a "find\_dosage" intent, and several entities are extracted, including information such as the duration and medicine/drug type. 

\section{Methods}

\subsection{Data Description \& Pre-Processing}
For extracting information from scientific literature related to COVID-19, we focused on CORD-19\cite{wang_cord-19_2020} (COVID-19 Open Research Dataset), which is an open-source, publicly available dataset of scholarly articles related to COVID-19, SARS-CoV-2, and other coronaviruses for use by the global research community. It is organized and updated by the Allen Institute daily, and as of September 11, 2020, CORD-19 has collected 253,454 publications. Supplementary datasets found on the COVID-19 Data Index (https://www.covid19dataindex.org/), such as the Symptoms Checker, were used to query the full text and help provide associated terms and medical information to use as a chatbot response.

In order to implement the NLP model to perform NER for drugs, symptoms, and other terms from the scholarly articles, we transformed the text from the CORD-19 dataset (titles, abstracts, and full-text JSONs from the collection of JSON files that contain full-text parse of a subset of CORD-19 papers) into a database, and then into a usable format (String) in Python.



\subsection*{Exploratory Analysis}
\begin{figure}[t!]
    \centering
    \includegraphics[width=0.5\linewidth,height=0.2\textheight]{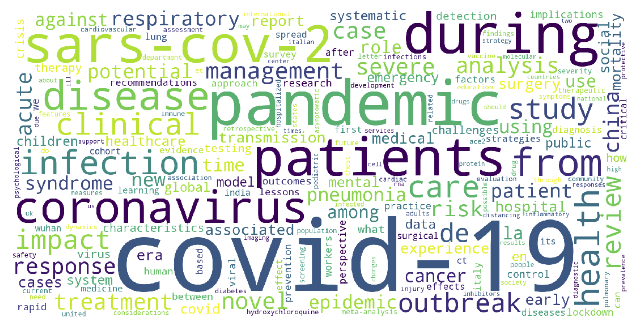}
    \caption{Most frequent words in titles of COVID-19 papers}
    \label{fig:word_cloud}
\end{figure}

We created a word cloud of the most frequent words within the titles of articles directly related to COVID-19 (Figure~\ref{fig:word_cloud}). From the word cloud, we identified words that show up more frequently than others, giving us insight into what words we would want to be included as nodes in our knowledge graph and whether or not they were captured as entities from NER.

\begin{figure*}[t]
    \centering
    \begin{subfigure}[t]{\linewidth}
        \centering
        \includegraphics[width=\linewidth, trim = 10 0 0 50, clip]{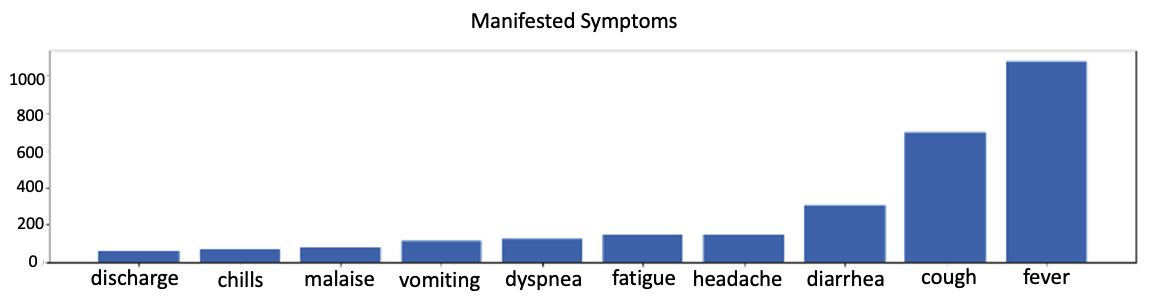}
        \vspace{-15pt}
        \caption{Frequency of COVID-19 symptoms in literature}
        \label{fig:freq_word}
        \vspace{15pt}
    \end{subfigure}
    
    \begin{subfigure}[t]{\linewidth}
        \centering
        \includegraphics[width=1.0\linewidth, trim = 0 0 0 30, clip]{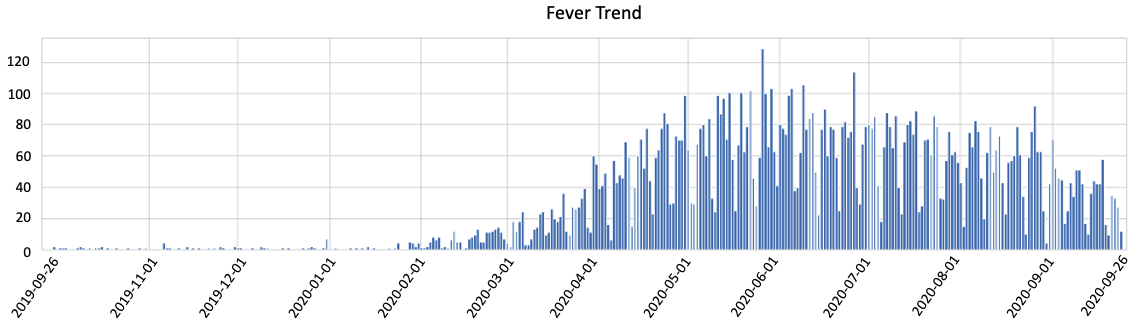}
        \vspace{-15pt}
        \caption{Trend of the frequency of ``fever''}
        \label{fig:fever}
        \vspace{5pt}
    \end{subfigure}
	\caption{Frequency of symptoms appearing in literature}
    \label{fig:frequency_symptoms}
\end{figure*}

Using the spaCy matcher library and a predefined list of generic symptoms from scientific literature, we did a basic exploration of the articles in the dataset. We counted the number of articles that reference certain symptoms related to viruses in the texts that have a reference to COVID-19. For example, the frequencies of the different symptoms appearing in literature are shown in Figure~\ref{fig:freq_word}. Of the most frequently mentioned symptoms, we can see that fever, cough, and diarrhea are the top three. We recorded how the frequency of the word ``fever'' increased over time as more papers on COVID-19 started to be published around March 2020~\ref{fig:fever}. Although symptoms like fever (Fig. \ref{fig:fever}), cough, and diarrhea have different frequencies, which is consistent with Fig. \ref{fig:freq_word}, they show a similar trend.

We can also see that there are papers about these symptoms before the start of the pandemic. We considered papers that are related to 'covid-19', 'coronavirus', 'cov-2', 'sars-cov-2', 'sars-cov', 'hcov', and '2019-ncov'. The papers before the pandemic examine the known coronaviruses and SARS cases. While fever, cough, and diarrhea are also symptoms of SARS, there are new symptoms that we discovered from this pandemic, namely loss of smell (anosmia) and loss of taste (ageusia). 


\subsection{Text Preprocessing and Extraction} \label{section:text_preprocessing}
\begin{figure*}[t]
    \centering
    \begin{subfigure}[t]{\linewidth}
        \centering
        \includegraphics[width=0.85\linewidth, height=0.35\textheight, trim = 0 0 0 50, clip]{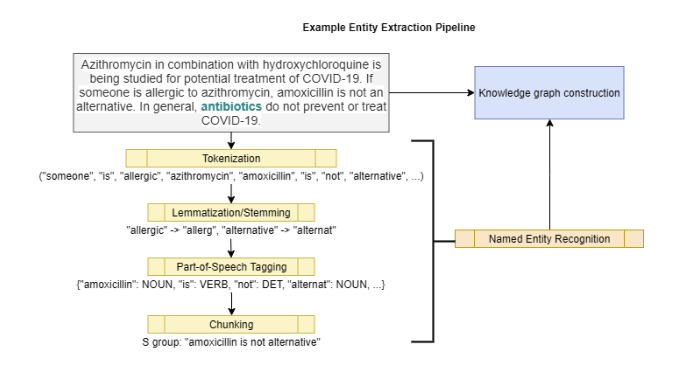}
        \vspace{-15pt}
        \caption{Entity Extraction pipeline given a piece of text from literature, which lists the various techniques that are behind NER}
        \label{fig:NER}
        \vspace{15pt}
    \end{subfigure}
    
    \begin{subfigure}[t]{\linewidth}
        \centering
        \includegraphics[width=.75\linewidth]{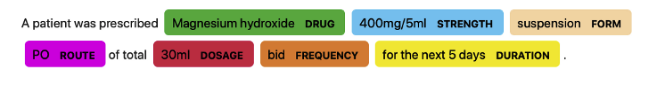}
        \vspace{-15pt}
        \caption{Med7 NER classification of words into 7 pre-defined labels}
        \label{fig:med7}
        \vspace{5pt}
    \end{subfigure}
	\caption{An example NER pipeline and example results from Med7 NER}
    \label{fig:NER_pipeline_example}
\end{figure*}

To understand the CORD-19 dataset, the specific feature engineering techniques and methods that we incorporated were tokenization, lemmatization, and, lastly, named-entity recognition (NER). We first tokenized the raw literature text data from our CORD-19 dataset for both the abstracts and full text. We removed stop words and other punctuation as the initial preprocessing steps and also performed lemmatization in order to extract root forms of words for more accurate tagging. An overview of these processes is shown in Figure~\ref{fig:NER}.

\begin{figure*}[t]
    \centering
    \subcaptionbox[]{Example knowledge graph\label{fig:knowledge_graph}}[.4\linewidth]{\includegraphics[width=0.9\linewidth,height=0.2\textheight, trim = 0 0 0 30,clip]{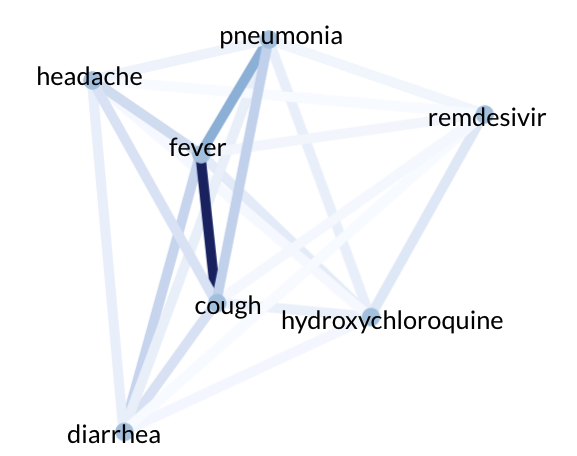}}
	\subcaptionbox[]{Patient subgraph\label{fig:trajectory}}[.49\linewidth]{\includegraphics[width=\linewidth,height=0.22\textheight]{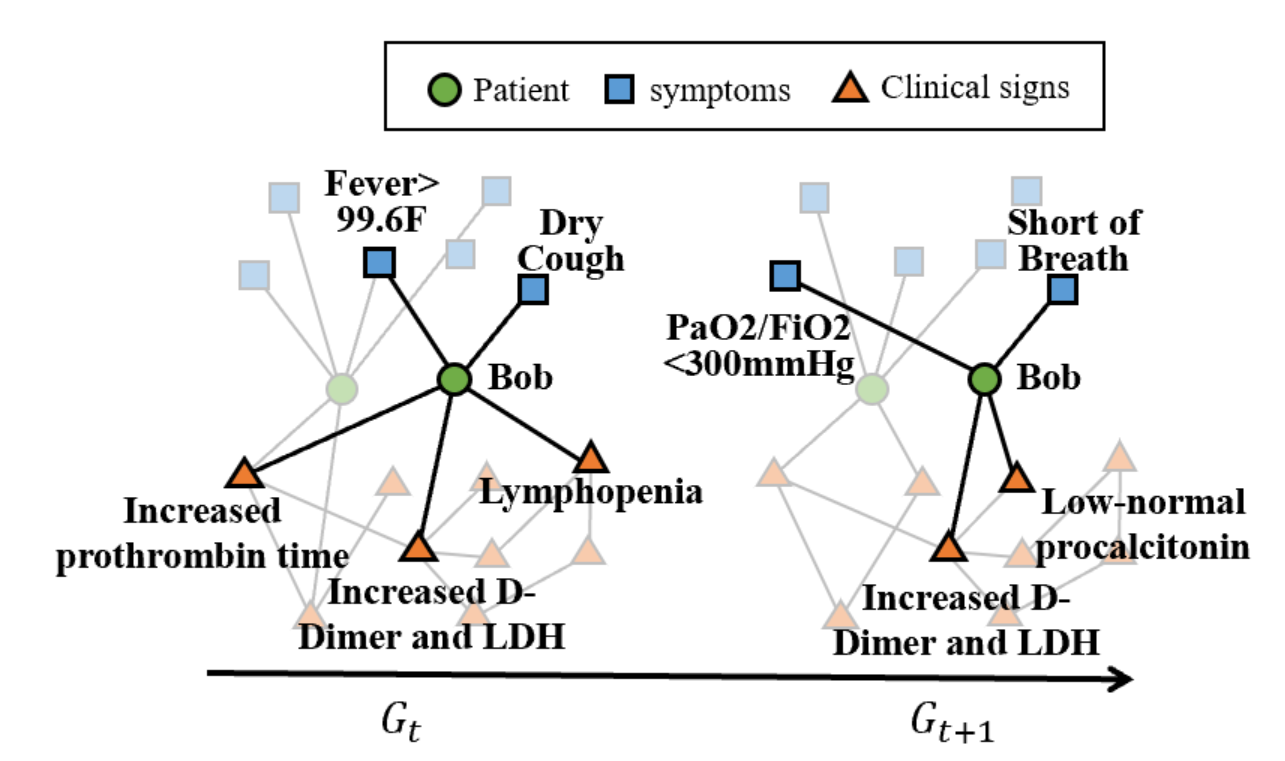}}
	\caption{A sample knowledge graph created using sciSpacy NER model and the evolution of a patient subgraph over time.}
    \label{fig:knowledgeGraph_patientSubgraph}
\end{figure*}

After performing NER on the abstracts and full text in the dataset, sentence-by-sentence associations between the symptoms were employed to build a knowledge graph. Building the knowledge graph is one of the major components of our proposed method. In addition to the information from NER, we incorporated natural language understanding (NLU) to extract features from chats using the chatbot to update the knowledge graph. The example of a knowledge graph is shown in Figure~\ref{fig:knowledge_graph}.

\subsection{Our Models} 
For this work, we used machine learning-based NLP techniques for interacting with patients to collect data and extract meaningful and useful information from it. Regular virtual appointments with the chatbot can help patients who are self-isolating or under-investigation for infection keep in touch with their healthcare provider. The chatbot can monitor symptoms and health conditions, and build a personalized patient profile, which allows providers to make decisions based on evidence.

The smartbot is trained using text data from the existing scientific research literature on COVID-19 and related diseases to learn the names of symptoms, drugs, and other terminology associated with the disease. Using this trained model, the smartbot is able to create a conversational environment to interact with the patients, answer questions from patients, and conduct in-depth exploration for concerning conditions. Our end-to-end chatbot is built on top of state-of-the-art framework Rasa \cite{bocklisch_rasa_2017}.

With the patient information retrieved from smart chatbot conversations, we built a personalized profile for each patient, recording symptom development and drug usage in a time series. Beyond the profile, we constructed a graph for each patient, one sub-graph for each conversation. The patient acts as the central node, with neighboring symptoms nodes (from the patient profile). We then map each sub-graph to the knowledge graph built from NER results in chronological order to get a patient-specific trajectory or a set of exterior symptom/condition nodes from the original knowledge graph that we can predict for the patient. In Figure \ref{fig:trajectory}, the patient subgraph is in bold while the translucent nodes along the fringe of the patient's subgraph form the patient's trajectory over time.

\subsection{NER Methods}\label{section:ner}
Since training NER models to fit COVID-specific categories require annotation, which is a collection of pre-tagged texts, it is infeasible to start from scratch using the full literature text. However, pre-trained, medical-related NER models, such as Med7~\cite{kormilitzin_med7_2020}, work well as an alternative. Med7 is a clinical NER that classifies tokens into seven categories which are sufficient for our use cases. In Figure~\ref{fig:med7}, when a patient provides a query asking for the duration that corresponds to Magnesium Hydroxide prescription (assuming he/she follows the conditions in which this sentence derives from) the trained knowledge graph can be searched to find the correct duration (5 days).

In the case that Med7 does not provide us sufficient data or features to answer basic queries, other medical NER models were explored. In our chatbot prototype, we combined Med7 and scispaCy~\cite{scispacy} to improve the entities extracted and model performance.

\subsection{Knowledge Graph}\label{section:knowledge_graph}
As we conducted NER on the dataset, we constructed a knowledge graph. Each named entity that we detect becomes a node in our Knowledge Graph. Rather than using edges to show a simple relationship or a physician’s decision, we weighted the edges to represent the significance of the relationship between two entities. These edge weights can be assigned by counting the number of times two named entities appear in the sentence by calculating conditional probabilities, especially for conditions and symptoms. This proves useful when making predictions or suggestions to the patients. If a patient has certain underlying conditions as well as symptoms, we use the edge weights to try and predict the likelihood that they have the virus. An entirely different method for assigning edge weights is to have a semantic meaning encoded into the edge instead. In the process of performing NER, we apply speech tagging to recognize the subject, object, and root verb of each sentence. The subject and object become nodes in the Knowledge Graph, and the root verb is used as a semantic descriptor on edge. This way, the sentence “A headache is a symptom of COVID-19” would make an edge from node “headache” to node “COVID-19” with the descriptor “symptom” in the Knowledge Graph.

\section{Results}
\subsection{Chatbot On Top of Rasa}\label{section:rasa} 

\begin{figure*}[t]
    \centering
    \subcaptionbox[]{A demo conversation between a patient and the chatbot in shell environment\label{fig:rasa_demo}}[.48\linewidth]{\includegraphics[width=.95\linewidth, height=0.2\textheight]{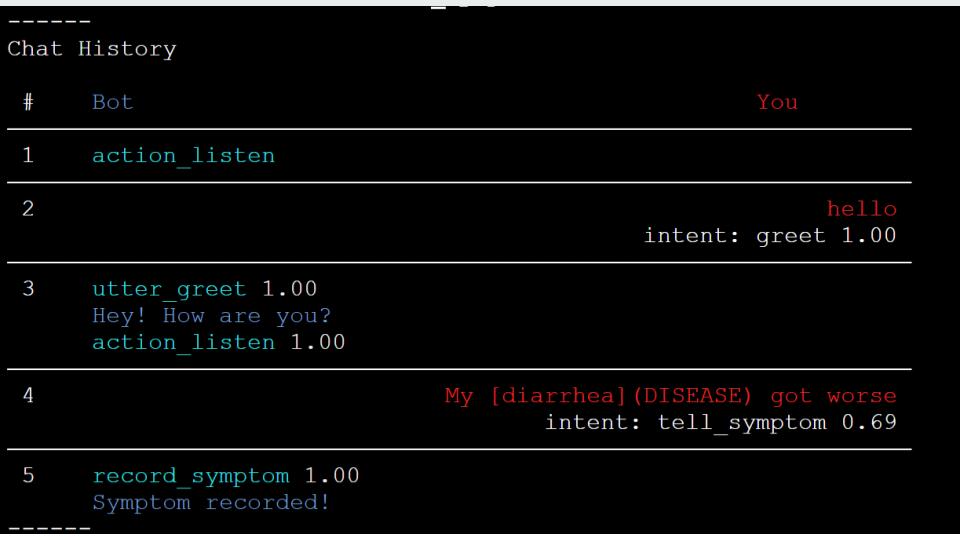}}
	\subcaptionbox[]{Example patient profiles\label{fig:patient_profile}}[.48\linewidth]{\includegraphics[width=.95\linewidth,height=0.18\textheight]{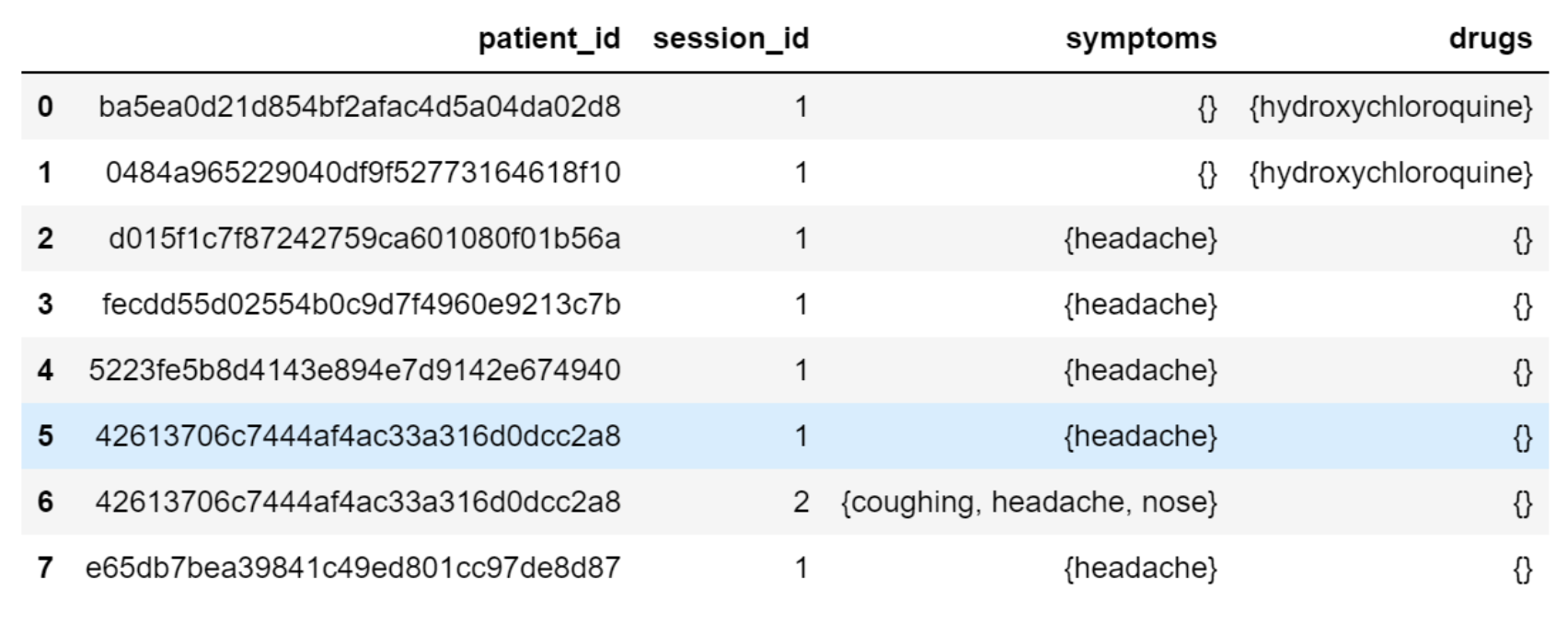}}
	\caption{A demo conversation showing actions of the chatbot and examples of patient profiles created. Each patient has a unique patient\_id. If there is no user activity for 1 hour, a new session will be started at the next activity. Symptoms are drugs are entities recognized from conversation history.}
    \label{fig:demo_profile}
\end{figure*}

For building our chatbot, we used the Rasa NLU framework, which also has built-in connectors with Facebook Messenger, Slack, Google Hangouts, Telegram, and several other popular social media applications\cite{bocklisch_rasa_2017}. Rasa met our requirements most readily because of its built-in entity extraction feature. We embedded pre-trained named entity recognition models, including those we used in Section \ref{section:ner} into the chatbot. This way, the chatbot is able to recognize entities never seen in training data. The language model we chose is the "en\_ner\_bc5cdr\_md" model, which is trained on a medical corpus, from scispaCy \cite{scispacy} as the built-in entity extractor and trained our Rasa model to recognize symptoms (marked as "DISEASE") and drugs (marked as "CHEMICAL" from patient input and record it as shown in Fig.\ref{fig:rasa_demo}. The bot message is on the left and the patient ("You") is on the right.  To make the conversation smooth, we also have basic conversation functions, including greeting, farewell, confirmation, and comforting (see Figure~\ref{fig:demo}).

Once we have collected a large number of patient trajectories, we can predict one patient's future trajectory if the current trajectory is similar to some existing trajectory on file. In other words, Bob in Figure \ref{fig:trajectory} acts as a "hidden" node, and through time, if other patients have experienced a similar set of exterior symptoms/nodes, then we can predict that Bob may have symptoms that develop from time $G_t$ to $G_{t+1}$, such as "Low-normal procalcitonin".

We then implemented a feature to recognize symptoms and drugs and respond with a corresponding "Symptom recorded" or "Drug recorded" message. First, any entity of "DISEASE" and "CHEMICAL" extracted by the scispaCy model is recognized as symptoms and drugs respectively. Additionally, we added descriptions of symptoms frequently related to COVID-19 in oral language as training examples so that the DIETClassifier\cite{rasa-diet} in Rasa can recognize them in causal communication. 
Given the patient profile database built as described in Section \ref{section:patientgraph}, we would also be able to raise personalized questions based on patient history. As shown in Figure~\ref{fig:followup}, if the latest state for one patient included the symptoms of nose, coughing, and headache, our chatbot would ask, "You mentioned in our last conversation that you have symptoms of nose, coughing, headache. How do you feel about them now?"

\subsection{Patient Information Retrieval and Further Analysis} \label{section:patientgraph}
Based on the input from the patient, we can extract entities including symptoms, drug usage, etc., to build a personalized profile for the patient. One example is shown in Fig. \ref{fig:patient_profile}. We match our patient’s responses, symptoms, conditions, and current state to nodes in the existing knowledge graph from Phase 1. Then, each patient’s state becomes a subgraph of the larger knowledge graph. Ultimately, we continue to build on top of this knowledge graph and formulate equivalent responses to be used on patients with similar symptoms or conditions.

\section{Discussion} \label{section:vaccine_discussion}
In this work, we have shown how NER methods can be applied to extract useful information on symptoms, medication, lab results, and other relevant factors from existing literature on COVID-19 and, in general, for any disease. This information can be fed into a chatbot framework for gathering health information from patients and general users. This framework provides an inexpensive and simple means for the public to be in contact with their healthcare providers even if in-person visits are not possible.

The chatbot can be an effective tool in monitoring the health of the users with minimal intervention from healthcare providers, thus saving on the time required by healthcare providers to keep in touch with every patient individually. Additionally, the chatbot can alert either the patient or the healthcare providers or both if the symptoms are getting worse over time and if intervention by caregivers is necessary. 

Further, the development of the knowledge subgraph for each patient allows a simple and effective tool for caregivers to understand the evolution of symptoms for a particular patient over time. Additionally, if data is collected from a sufficiently large patient cohort over time, the patient subgraphs and their trajectories over time can be ``mined'' to discover useful information about the progression of a disease. This mechanism could also be used for predicting which symptoms a patient is likely to develop over time by comparing the patient's subgraph trajectory with other similar trajectories in the database. As a result, it might be possible to prepare for or prevent severe symptoms beforehand, thus improving patient outcomes.

As vaccines and other strategies for the prevention and treatment of COVID-19 are discovered over time, the chatbot can become an effective tool in collecting and managing data with regard to the conditions of a person even after the vaccine is administered. Results from such studies are likely to be extremely useful in determining vaccine efficacy and the need for booster vaccines in the future. 

Moreover, since the healthcare system in several countries is being overwhelmed due to an excessive number of cases, conducting follow-ups with patients after they have been discharged from the hospital is often a challenge. The chatbot can be used by hospitals for conducting follow-up conversations with patients after they have been released subsequent to being treated for COVID-19. This will aid not only in monitoring patients with minimal effort from physicians, nurses, and other healthcare personnel but will also be useful in determining if there are re-infections in past patients.

Lastly, our proposed methods can be used effectively for diseases other than COVID-19 to monitor symptoms and conduct ``virtual appointments'' with the chatbot, so that symptoms are monitored and recorded much quicker than actual in-person consultations. This is especially helpful for diseases that are time-sensitive in nature, such as stroke and cardiac arrest. In these scenarios, since past patients are likely to have a subsequent stroke or cardiac arrest, monitoring their health through regular chatbot conversations might prove invaluable in prevention and timely intervention, thus saving lives and improving outcomes in patients.

\section{Conclusion} \label{section:conclusion}
We demonstrated a smart chatbot prototype in our project to help monitor patients during the pandemic. The model can be generalized to handle other types of patient follow-ups and long-term monitoring. The chatbot can interact with individuals in a personalized manner to build digital profiles about symptoms and conditions, which can enrich electronic healthcare records to better inform providers in early intervention and treatment.



\end{document}